%% file: main.tex
\documentclass[sigconf]{acmart}

\AtBeginDocument{%
  }

\setcopyright{acmcopyright}
\copyrightyear{2022}
\acmYear{2022}
\acmDOI{}

\acmConference[ANDEA’22]{The 2nd KDD Workshop on
Anomaly and Novelty Detection, Explanation and Accommodation}{August 14,
  2022}{Washington, DC}
\acmPrice{}
\acmISBN{}

\usepackage{tikz}
\usepackage{pgfplots}
\usepackage{cleveref}
\usepackage{caption}
\usepackage{subcaption}
\usepackage{graphicx}
\usepackage{geometry}
\begin{document}

\title{Prototypes as Explanation for Time Series Anomaly Detection}

\author{Bin Li}
\email{bin.li@tu-dortmund.de}
\orcid{0000-0002-9707-4596}
\affiliation{
  \institution{TU Dortmund University}
  \city{Dortmund}
  \country{Germany}
}
\author{Carsten Jentsch}
\email{jentsch@statistik.tu-dortmund.de}
\affiliation{
  \institution{TU Dortmund University}
  \city{Dortmund}
  \country{Germany}
}

\author{Emmanuel M{\"u}ller}
\email{emmanuel.mueller@tu-dortmund.de}
\orcid{0000-0002-5409-6875}
\affiliation{%
  \institution{TU Dortmund University}
  \city{Dortmund}
  \country{Germany}}

\begin{abstract}
Detecting abnormal patterns that deviate from a certain regular repeating pattern in time series is essential in many big data applications. However, the lack of labels, the dynamic nature of time series data, and unforeseeable abnormal behaviors make the detection process challenging. Despite the success of recent deep anomaly detection approaches, the mystical mechanisms in such black-box models have become a new challenge in safety-critical applications. The lack of model transparency and prediction reliability hinders further breakthroughs in such domains. This paper proposes ProtoAD, using prototypes as the example-based explanation for the state of regular patterns during anomaly detection. Without significant impact on the detection performance, prototypes shed light on the deep black-box models and provide intuitive understanding for domain experts and stakeholders. We extend the widely used prototype learning in classification problems into anomaly detection. By visualizing both the latent space and input space prototypes, we intuitively demonstrate how regular data are modeled and why specific patterns are considered abnormal.
\end{abstract}




\maketitle

\section{Introduction}

Anomaly detection in time series data is gaining traction in several current big data research areas. The vast development of data processing and analysis facilitates real-time data monitoring applications in different branches, e.g., health care, manufacturing, astronomy \cite{hundman2018detecting, ren2019time, ming2019interpretable}. Anomaly detection in time series data is an important application area in those scenarios. 
Unlike traditional outlier detection in the database system, detecting single abnormal points or even whole abnormal time periods in time series data is much more challenging due to its serial dependent and dynamic nature. On the one hand, time series data is usually collected from sensor networks in real-time, leaving less time for domain experts to generate labels for the model training. On the other hand, anomalies often only appear on specific dimensions (subspace anomalies) and specific temporal contexts (contextual anomalies), making detecting and interpreting such anomalies even harder. 
\par
In recent years, deep reconstruction-based approaches have been developed for time series anomaly detection tasks, for example, the autoencoders \cite{malhotra2016lstm, zong2018deep}. An autoencoder is a symmetric neural network trained to reconstruct the regular\footnote{To prevent ambiguity, normal data in the anomaly detection context is referred to as regular data in this paper.} data and is supposed to produce significantly larger reconstruction errors by unseen anomalies. Specifically for time series data, recurrent neural networks (RNNs) are used to construct the autoencoders to capture the temporal information in the input sequences. Thanks to their model capability, deep autoencoders have shown convincing performance in detecting high dimensional and temporal anomalies from time series \cite{malhotra2016lstm}. 
\par 
Despite the performance advantage, autoencoders also suffer the criticisms of other deep neural networks - the lack of transparency. Though the detected anomalies fit well with standard evaluation criteria, their reliability is not well-proven. The lack of human interpretable information makes it hard to tell why an anomaly is abnormal, especially in high-dimensional time series data or long input sequences. In safety-crucial applications, for instance, a machine learning model assists in detecting abnormal arrhythmia in patient ECG data. Without a verifiable interpretation, it might lead to catastrophic consequences by automatic treatment. Therefore, interpretability is an essential demand of such anomaly detection applications.
\par
To address the black-box issue, various interpretable machine learning approaches recently gained attention in the fields that require high transparency and human-understandability \cite{schlegel2020empirical}. We propose to use example-based prototypes as an intuitive and explainable solution for interpreting anomalies in time series data. Prototypes are widely used for case-based reasoning in computer vision \cite{li2018deep, chen2018looks, hase2019interpretable}, graph learning \cite{zhang2021protgnn} and sequential data learning \cite{ni2021interpreting, ming2019interpretable, gee2019explaining}. Embedding a prototype layer is one of the widely used approaches to learn prototypes from neural networks in an end-to-end fashion. With the build-in prototype layer in the neural networks, prototype-based models are commonly efficient to train without requiring extra investigation in the interpretability functionalities. Moreover, the prototypes are usually self-contained and straightforward to understand, e.g., representative animal faces, sentences, and sensor data patterns. However, prototypes are still understudied in the anomaly detection field. 
\par 
In this paper, we propose to use prototypes to interpret the regular data during anomaly detection using autoencoders. In this context, data showing a certain repeating regular pattern is generated by several latent distributions, while the anomaly is any data point or period that deviates from from this regular pattern. We model the regular patterns of the time series data with prototypes in the latent space of the autoencoder and learn multiple prototypes to discover the latent components of the regular data distribution. Anomaly patterns that lie distantly from the data generated during the regular pattern state can be explained by comparing them with the constructed prototypes.
\par
Moreover, we further explain the anomaly patterns that lie distantly from regular data in the latent space by comparing them with the prototypes. To our knowledge, this is the first prototype-based explanation application in the time series anomaly detection domain. 
\par
Our contribution to the paper can be summarized as follows:
\begin{enumerate}
    \item we propose ProtoAD, an end-to-end LSTM-Autoencoder for anomaly detection with prototype learning
    \item we develop latent space prototype-based explanations for the understanding of the regular state of the studied data
    \item we evaluate our method with synthetic and real-world time series data. Moreover, we visually demonstrate examples of prototypes to show the benefit of our model qualitatively
\end{enumerate}

\section{Related works}
This section briefly reviews the existing work in autoencoder-based anomaly detection models and prototype-based explanations methods.
\subsection{Reconstruction-based anomaly detection}
Autoencoders have been used as an unsupervised anomaly detection approach for years. 
Feed-forward autoencoders \cite{zong2018deep} and Variational Autoencoders \cite{xu2018unsupervised} are used for time-independent data. In contrast, RNN-based autoencoders \cite{malhotra2016lstm} show their strength in detecting contextual anomalies in time series data. Based on the reconstruction error, a standard approach for estimating anomaly likelihood is to assume the reconstruction error following a normal distribution and measure the Mahalanobis distance between the reconstruction error of unknown data and the estimated distribution \cite{malhotra2016lstm}. In addition to reconstruction error, the hidden representation in the latent space can also be used for likelihood estimation  \cite{zong2018deep}. Gaussian Mixture Model (GMM) \cite{zong2018deep} and energy-based model \cite{zhai2016deep} are also used for the likelihood estimation. Common thresholding techniques over the anomaly likelihood are based on maximizing the performance on a validation set, which requires labels in advance \cite{malhotra2016lstm}. 
\par
Other approaches, including the hierarchical temporal memory (HTM) \cite{ahmad2017unsupervised} and temporal convolutional network (TCN) \cite{he2019temporal} are also adopted in time series anomaly detection concerning different use cases and data properties. However, they are not directly relevant to the reconstruction-based models. 

\subsection{Explanation with prototypes}

Due to the complex properties of both feature and time dimensions of time series data, prototypes are considered an intuitive explanation. Common prototype learning approaches for neural networks follow a three-step paradigm. 1) Representation learning, 2) prototype learning in the latent space, and 3) class prediction. The objective commonly includes 1) minimizing classification error, 2) minimizing the distances between each hidden representation and one of the prototypes, and 3) maximizing the distances between prototypes. 
\par
In the existing prototype learning literature, \cite{li2018deep} employs a multi-layer convolutional neural network to construct the autoencoder, which learns hidden representations for image data. They rely on the decoder to project the learned prototypes in the human-understandable space, sometimes producing unrealistic reconstructions. Using a single encoder to replace the autoencoder is considered as a reduction of training effort in  \cite{chen2018looks,ming2019interpretable}, and they use the nearest neighborhood of each prototype in the latent space as the corresponding realistic patterns in the input space. Under different problem settings, \cite{chen2018looks} and \cite{hase2019interpretable} build up the encoder with convolutional neural networks to encode image data, \cite{ming2019interpretable} uses RNNs for sequential data,  \cite{ni2021interpreting} use a convolutional layer to learn time series representations, \cite{zhang2021protgnn} employs graph neural networks for the encoder. In our work, we use the single LSTM-Autoencoder for both reconstruction-based time series anomaly detection and hidden space representation learning. 
\par
The standard objective functions of existing prototype learning approaches consist of multiple regularisation terms that are trained jointly. To ensure the representation ability of the prototypes,  \cite{li2018deep,ming2019interpretable,gee2019explaining,zhang2021protgnn,chen2018looks} all minimize the distance between each prototype and every nearby hidden representation as well as every hidden representation to each prototype. Furthermore, the learned prototypes are supposed to be diverse from each other \cite{ming2019interpretable, zhang2021protgnn,gee2019explaining}. In the objective, we will follow the standard design of the regularization terms above. However, different from most existing works, which use cross-entropy for their classification task to minimize the classification error \cite{li2018deep,ming2019interpretable,hase2019interpretable}, in our unsupervised setting, we use the reconstruction-error to regularize the reconstruction process of regular data. 
\par
Besides prototypes, other techniques are also used for explaining time series data. The representative subsequences Shapelets \cite{kidger2020generalised, li2020efficient} can be similarly used for explanation.  Instead of finding the representative pattern as prototypes, counterfactuals \cite{ates2021counterfactual} explain the instance towards the opposite class. Recently, the attention mechanism is also used for explaining time series data \cite{li2020efficient, vinayavekhin2018focusing}.

\section{Preliminaries}
\subsection{Terminology}

Let $X=\{X_t\}_{t\in\mathbb{Z}}$ be a $d$-dimensional time series process that shows a regularly repeating pattern over time periods of some length $L$. These repeating patterns are contained in sliding windows  $W_t=\{X_{t+1},\ldots,X_{t+L}\}$ of $L$ consecutive elements of the time series. Often, the window size $L$ can be selected based on prior knowledge on the dataset, which is known to show seasonality over e.g., one day or one week. 
\par
Anomalies in time series data are commonly divided into three categories \cite{choi2021deep}, point anomaly, contextual anomaly, and collective anomaly. In this work, we consider point anomalies (e.g., abrupt peaks in the data window) and contextual anomalies (e.g., the appearance of one or more points makes a temporal context unusual). Formally, we assume that the data points are generated by a periodically stationary time series process $X$ with periodicity $L$ \cite{wang2015detecting, ursu2009modelling}. That is, the time series consists of regularly repeating patterns of length $L$ which evolve over time without distributional changes, i.e., we do not consider concept drifts \cite{gama2014survey}. 
\par
Let  ${(W_t, y_t)}_{t\in\mathbb{Z}}$ be the dataset after applying the sliding window and $y_i\in\{0,1\}$ is the label of the window $W_t$ ($0$ for regular data and $1$ for anomaly). The anomaly detection is conducted on the window level. A window is considered abnormal if at least one point or sub-window with multiple points that shows a significantly different behavior to the window during the regular pattern state. The significance is determined by comparing the window anomaly score predicted by the model and a user-defined threshold.

\subsection{Problem definition}
Given the multi-dimensional time series data with applied sliding window, the target is to train an autoencoder-based end-to-end anomaly detector that 
\begin{enumerate}
    \item detect anomaly windows in an unsupervised manner
    \item deliver representative prototypes of regular data in the latent space
    \item leverage interpretation of anomalies based on the prototypes of regular data
\end{enumerate}

\section{Methodology}
In this section, we propose ProtoAD, an LSTM-Autoencoder with an additional prototype layer, which can be trained end-to-end in an unsupervised manner.  
\subsection{ProtoAD architecture}
The architecture of ProtoAD is in line with the existing prototype neural networks  \cite{ming2019interpretable,li2018deep}. We use an LSTM-Autoencoder to learn time series hidden representations in the latent space and feed the representations to the prototype layer for similarity-based prototype comparison. Specifically, we designed the architecture and training procedure for unsupervised anomaly detection, while only data consisting of regularly repeating patterns is used for the training and prototype learning. An overview of the ProtoAD architecture is shown in \autoref{fig:overview}. 
\par
\input{flowchart_new}

We construct the LSTM-Autoencoder in the fashion of \cite{malhotra2016lstm}. More specifically, the $d$ dimensional input window $W_t=\{X_{t+1}^{t+L}\}$  is feed into the encoder
\[\textbf{f}\colon\mathbb{R}^{L\times d}\to\mathbb{R}^{m}\]
The last hidden state of the encoder LSTM unit $h_{i}=\textbf{f}(W_t)$ ($h_i\in\mathbb{R}^m$) is used as the hidden representation of the input window in latent space. A same-structured decoder 
\[\textbf{g}\colon\mathbb{R}^{m}\to\mathbb{R}^{L\times d}\] 
target at reconstructing the window from the hidden representation $W_t^{'}=\{X_{t+1}^{'t+L}\}$. The decoder LSTM unit takes $h_i$ as the initial hidden state while takes the real data from previous timestamp as input. We train the autoencoder to minimize the reconstruction error of regular windows, i.e., no anomaly data will be used during training. 
\par
The reconstruction error at timestamp $t$ is defined as 
\[e_t=|X_t-X_t^{'}|\] 
The training set is used to estimate a normal distribution $\mathcal{N}(\mu,\Sigma)$ ($\mathcal{N}(\mu,\sigma)$ for univariate data) of the reconstruction error for multivariate input data. And the likelihood of a data point being abnormal is defined by the anomaly score
\begin{equation*}
a_t = \begin{cases}
\frac{1}{{\sigma \sqrt {2\pi } }}e^{{{ - \left( {e_t - \mu } \right)^2 } \mathord{\left/ {\vphantom {{ - \left( {x - \mu } \right)^2 } {2\sigma ^2 }}} \right. \kern-\nulldelimiterspace} {2\sigma ^2 }}} &d=1\\
% e_t & d=1 \\
(e_t-\mu)^T{\Sigma}^{-1}(e_t-\mu) &d>1
\end{cases}
\end{equation*}
The largest anomaly score is picked up to represented the window anomaly score \[a_{t+1}^{t+L}=\underset{i=1,...,L}{max}(a_{t+i})\]
In our work, we do not specify a threshold over the window anomaly scores to get a binary prediction. Instead, we directly evaluate the AUC score based on the real-valued anomaly scores. Different existing thresholding techniques can be applied to get a binary prediction in such a situation \cite{malhotra2016lstm, su2019robust, hundman2018detecting}. 
\par
Based on the anomaly detection model above, we introduce a prototype layer between the encoder and decoder, which leverages interpretable prototypes of the regular data during the end-to-end training process. The prototype layer does not influence the information flow from the encoder to the decoder, i.e., the only information the decoder gets from the encoder is the last encoder hidden state. The prototypes are learned by regularizing the objective function.  
\par
The prototype layer contains $k$ prototypes $p_i\in\mathbb{R}^m (i=1...k)$ to be learned, where $k$ is a user-defined parameter and $k$ vectors are randomly initialized within the range $[-1,1]$.
As introduced in \autoref{sec:obj}, several regularization terms are employed in the objective function to get the expected prototypes. 
\par
In most existing prototype-based models for classification task \cite{ming2019interpretable,kim2016examples,gee2019explaining},  the prototype layer is followed by some linear layers and a Softmax layer for the production of prediction, which increases the complexity of the network and requires additional regularization to enforce interpretability. As an anomaly detection model, our outputs are derived directly from the autoencoder reconstruction errors. Therefore, we omit the other common output layers after the prototype layer to simplify the network structure.

\subsection{Objective function}
\label{sec:obj}

The objective in the training phase is to train the autoencoder with regular windows such that the reconstruction error is minimized and to learn a batch of prototypes from the regular data.
The reconstruction error loss of the autoencoder is given by 
\[\mathcal{L}_{e}=\frac{1}{n}\sum_{i=1}^{n}\sum_{l=1}^{L}e_{t+l}\]
where $n$ is the number of sliding windows. 
To ensure the learned prototypes are informative and diverse enough to each other, we use the diversity loss designed by \cite{ming2019interpretable},
\[\mathcal{L}_{d}=\sum_{i=1}^{k}\sum_{j=i+1}^{k}max(0, d_{min}-||p_i-p_j||_2^2)^2\]
which defines the threshold $d_{min}$ that only apply this penalize to nearby prototype pairs. 
Finally, to ensure the prototypes are representative to the local hidden representations, we define the following representation regularization term
\[\mathcal{L}_{r}=\frac{1}{k}\sum_{j=1}^{k}\underset{i\in[1,n]}{min}||p_j-h_i||^2 + \frac{1}{n}\sum_{i=1}^{n}\underset{j\in[1,k]}{min}||h_i-p_j||^2\]
The first term ensure that each prototype is close to at least one hidden representation, which the second term ensures each hidden representation has one prototype to be represented.
\par
The overall objective function is 
\[\mathcal{L}=\lambda_{e}\mathcal{L}_{e}+\lambda_{d}\mathcal{L}_{d}+\lambda_{r}\mathcal{L}_{r}\] where $\lambda_{e}$, $\lambda_{d}$ and $\lambda_{r}$ are weighting hyperparameters.

\section{Experiments}

In this section, we introduce the experiments on ProtoAD under different settings. We experiment over different real-world datasets with a variety of anomaly types. In addition, to evaluate the model performance on specific data characteristics, we also introduce a synthetic dataset with artificial anomalies. Finally, we demonstrate the prototypes visually and analyze the prototype properties w.r.t. a variety of parameter settings.  

\subsection{Experiment setup}
\subsubsection{Datasets} We experiment on one synthetic dataset and four common real-world benchmark datasets in the time series anomaly detection domain. The dataset properties are summarised in \autoref{tab:data}. 
\par
To understand the anomaly detection process and the learned prototypes, we introduce a one-dimensional synthetic dataset sampled from a sine wave with amplitude $1$ and period $100$ timestamps. A random noise $\epsilon\in[0, 0.1]$ is added to every timestamp. In addition, we add a random factor ${\alpha}\in[0,1]$ every $100$ timestamps in the test set to simulate point anomalies. We define a half period of the sine wave as the window length (i.e., $L=50$), such that the model is supposed to learn the crests and troughs as two types of prototypes. 
\par
The New York City Taxi (Taxi) dataset is a one-dimensional real-world dataset with a clear periodical feature. It recorded the passenger counts over days in 2014. Extreme passenger counts on public holidays are considered anomalies. Following \cite{cui2016comparative} we aggregate the count numbers into $30$-minute intervals. We take one day (i.e., $L=48$) as the window length. 
\par
SMAP (Soil Moisture Active Passive satellite) and MSL (Mars Science Laboratory rover) are two multivariate telemetry datasets from NASA \cite{hundman2018detecting}. The datasets contain both point and contextual anomalies. Domain experts labeled the test sets. However, there are also anomaly data in the training sets. The polluted training set can impact the purity of prototypes. There is no common repeating pattern in the datasets. We set the window length $L$ as $100$ for both datasets. 
\par
SMD (Server Machine Dataset) \cite{su2019robust} is a multivariate dataset collected from servers of an Internet company. The dataset is already divided into two equal-sized training and test sets by the provider and labeled anomalies by domain experts based on incident reports. We only use the data from one machine (machine-1-1) in our experiments. We set $L=100$ for SMD.

\begin{table}[htbp]
\caption{Datasets}
\begin{center}
\begin{tabular}{|c|c|c|c|}
\hline
& Length & Dimensionality & Anomaly rate (\%) \\ \hline
Synthetic & $20 000$ & $1$ & $1.0$\\
Taxi & $17 520$ & $1$ & $4.4$\\
% ECG5000 & 5 000 & 1 & 6.2\\ 
SMAP & $562 800$& $25$ & $13.1$\\ 
MSL & $132 046$ & $55$ & $10.7$ \\
SMD & $56 958$ & $38$ & $9.5$\\\hline
\end{tabular}
\label{tab:data}
\end{center}
\end{table}

\subsubsection{Evaluation metrics}
We adopt the AUC score as the evaluation metric. Considering the essential requirement of detecting both point and contextual anomalies, we only evaluate on the window level. A data window is abnormal if it contains one or multiple abnormal instance(s).

\subsubsection{Competitors}
To the best of our knowledge, this is the first work that engages time series anomaly detection and prototype learning. The existing prototype learning networks \cite{ming2019interpretable, ni2021interpreting, gee2019explaining} commonly work in a supervised manner, which requires labeled data for the training phase. Therefore they are not directly relevant to our setting. We mainly compare our method with the unsupervised anomaly detection approaches. Firstly, we compare with the EncDecAD \cite{malhotra2016lstm}, which has a similar setting as ours, but without the prototype layer. Thereby, we can determine whether the prototype learning damages the original reconstruction-based anomaly detection. Furthermore, we compare with one of the state-of-the-art unsupervised time series anomaly detection OmniAnomaly \cite{su2019robust}. We follow most of the default hyperparameter settings in \cite{su2019robust} but use the window length same as in our work for the sliding windowing. 
\subsubsection{Experimental details}
In all experiments, we set  ${\lambda}_e=0.025$, ${\lambda}_d=0.2$ and ${\lambda}_r=0.5$. During training, $25\%$ of the data is used for learning the parameters $\mu$ and $\Sigma$ ($\sigma$ for univariate data). All models are trained for $100$ epochs with batch size $20$, learning rate $0.0001$, dropout rate $0.2$. We use the Adam optimizer \cite{kingma2014adam}. All experiments are conducted on a NVIDIA Quadro RTX 6000 24GB GPU. The experimental results are averaged over three runs. 
\subsection{Evaluation results}
\subsubsection{Anomaly detection performance}
\begin{table}[htbp]
\caption{Anomaly detection performance}
\begin{center}
\begin{tabular}{|c|c|c|c|}
\hline
& EncDecAD & ProtoAD & OmniAnomaly \\ \hline
Synthetic & $0.50$ & $0.54$ & $0.95$\\
Taxi & $0.53$ & $0.63$ & $0.52$ \\
% ECG5000 & & & \\ 
SMAP & $0.41$ & $0.40$ & $0.49$\\ 
MSL & $0.73$ & $0.73$ & $0.50$ \\ 
SMD & $0.95$ & $0.95$ & $0.51$\\ \hline
\end{tabular}
\label{performance}
\end{center}
\end{table}

Firstly we report the AUC score over different models in \autoref{performance}. For ProtoAD, we take the number of prototypes $k=10$. There is no significant difference between EncDecAD and ProtoAD, which indicates that the additional prototype layer and corresponding learning process do not directly impact the anomaly detection performance. ProtoAD even benefits from the prototype learning in the Synthetic and Taxi datasets. OmniAnomaly shows worse AUC scores in comparison with the other two models. Different from \cite{su2019robust}, where all possible thresholds over the predicted anomaly scores are traversed, and only the threshold with the best $F1$ score is reported, the AUC score reflects more general quality of the anomaly scores over multiple thresholds. 

\subsubsection{Parameter sensitivity}
\begin{figure}[h]
\includegraphics[scale=0.5]{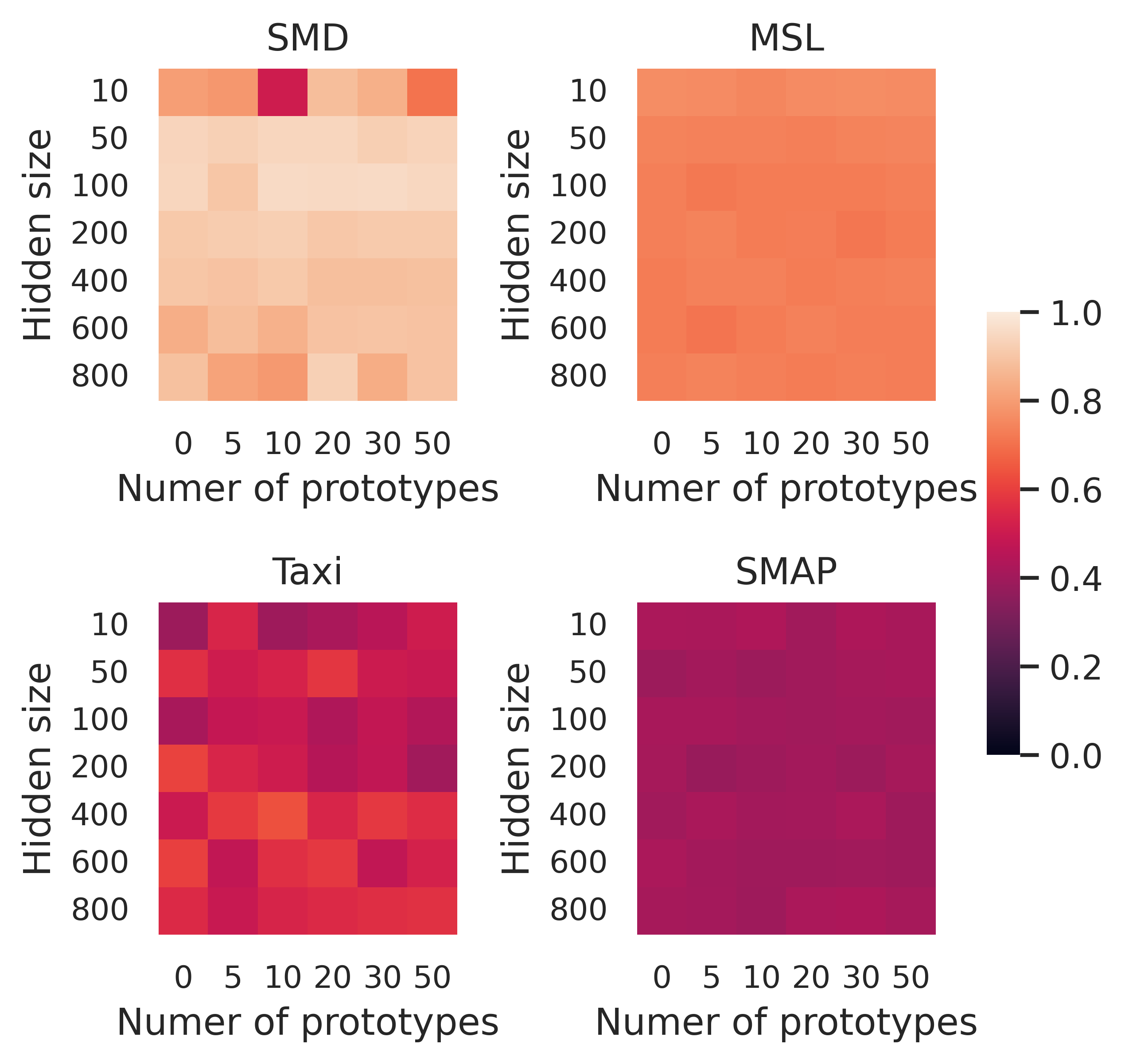}
\caption{AUC score under different parameter setting}
\label{parameter}
\end{figure}

The hidden layer size $m$ and the number of prototypes $k$ are two major hyperparameters in ProtoAD. In this section, we examine the performance sensitivity to those two parameters. For each dataset, we try $m\in[10, 50, 100, 200, 400, 600, 800]$ and $k\in[0, 5, 10, 20, 30, 50]$, where $k=0$ reduces ProtoAD to EncDecAD. Heatmaps of the AUC scores on real-world datasets are shown in \autoref{parameter}. The results indicate that the model is sensitive to neither hidden size nor the number of prototypes, as far as the hidden size is large enough to capture all information in the data windows. However, the number of prototype $k$ should not be set too large. Otherwise, the model tends to learn redundant prototypes (see \autoref{fig:umap}).

\subsubsection{Latent space visualization}
\begin{figure}[h]
\includegraphics[scale=0.25]{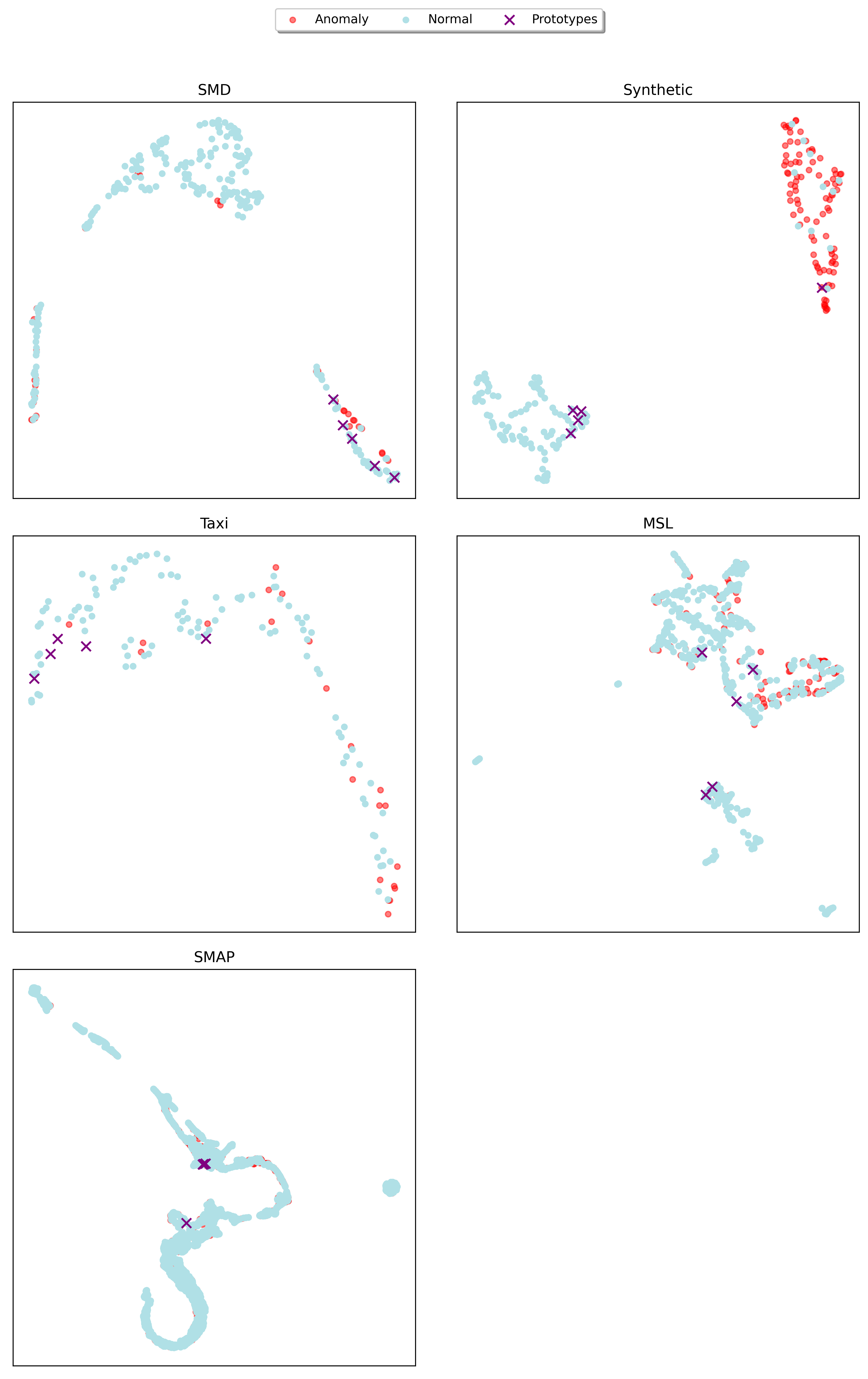}
\caption{The UMAP visualization of the ProtoAD latent space}
\label{fig:umap}
\end{figure}

We investigate a visualization of the autoencoder hidden space in this section to understand how time series data windows are embedded and how prototypes of regular data are learned. We use UMAP \cite{mcinnes2018umap} to reduce the high-dimensional latent representations into two dimensions. The result is visualized in \autoref{fig:umap}. Here, we set $k=5$ for all datasets. The prototypes shown in the plots are learned during the training phase. The plotted regular and anomaly points are embedded from the test data.
\par
In the synthetic data, the regular data lie in two regions. Four prototypes are learned from the trough half (lower left) and one from the crest half (upper right). Since the anomalies are always generated at the beginning of the crest half, the anomalies lie nearer to the crest regular data embedding. 
\par
In the real-world datasets, especially the SMAP and MSL with polluted training data, regular and abnormal data do not clearly show separated clusters, though the learned prototypes represent the major blocks of dense regions showing regular patterns. Specifically, the prototypes gather at the bottom right corner for SMD, while no prototype is at the larger upper cluster. A possible reason is that the high-dimensional server data contain many zero values. The model can not summarize informative patterns in the training set, and slightly different regular patterns in the test data are mapped into a different region.

\begin{figure}
     \centering
     \begin{subfigure}[b]{0.25\textwidth}
         \centering
           \includegraphics[scale=0.30]{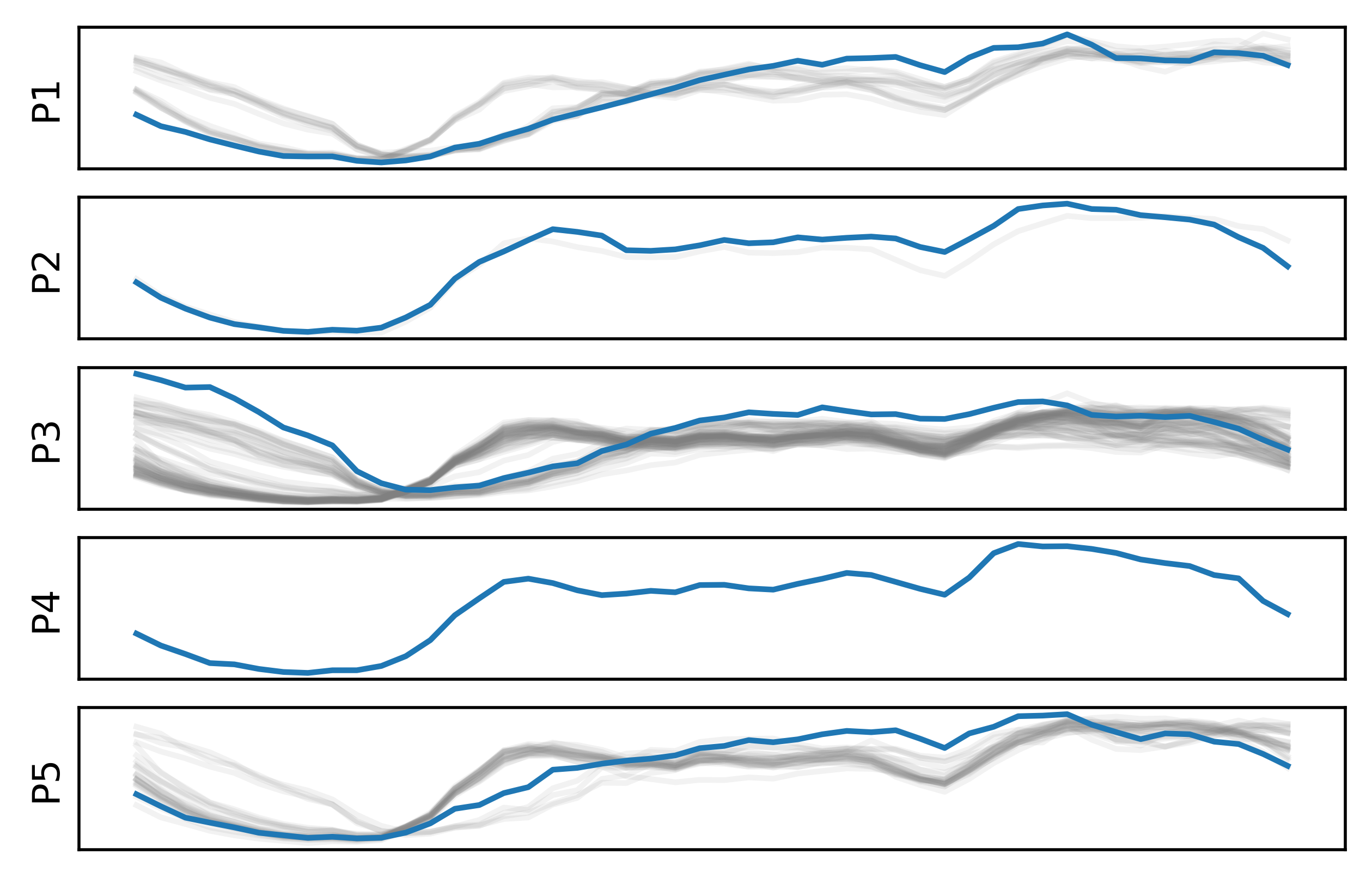}
         \caption{Taxi regular}
         \label{fig:taxi_n}
     \end{subfigure}
    %  \hfill
    ~
     \begin{subfigure}[b]{0.25\textwidth}
         \centering
         \includegraphics[scale=0.30]{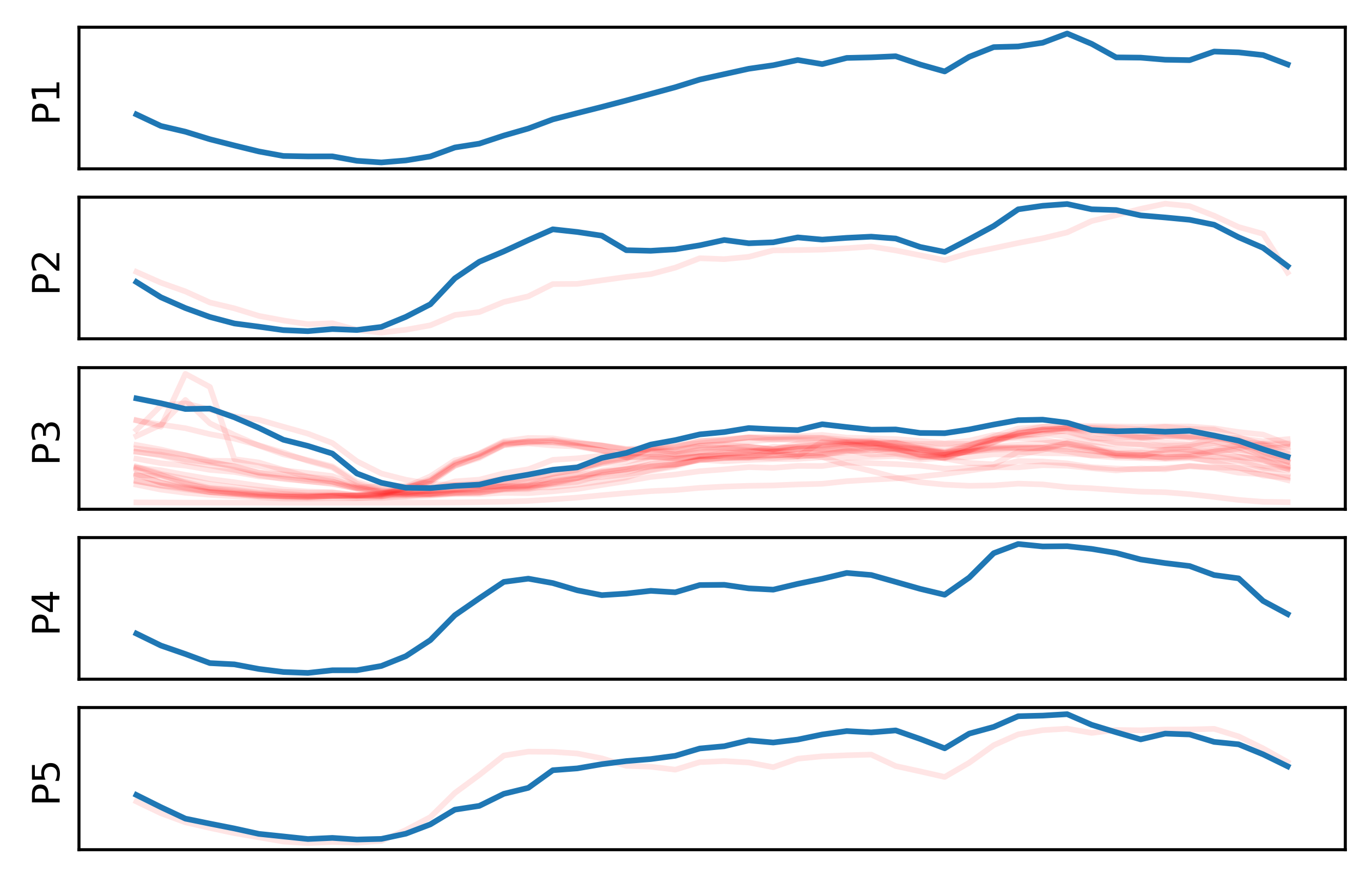}
         \caption{Taxi anomaly}
         \label{fig:taxi_a}
     \end{subfigure}
    %  \hfill
    
     \begin{subfigure}[b]{0.5\textwidth}
         \centering
        \includegraphics[scale=0.50]{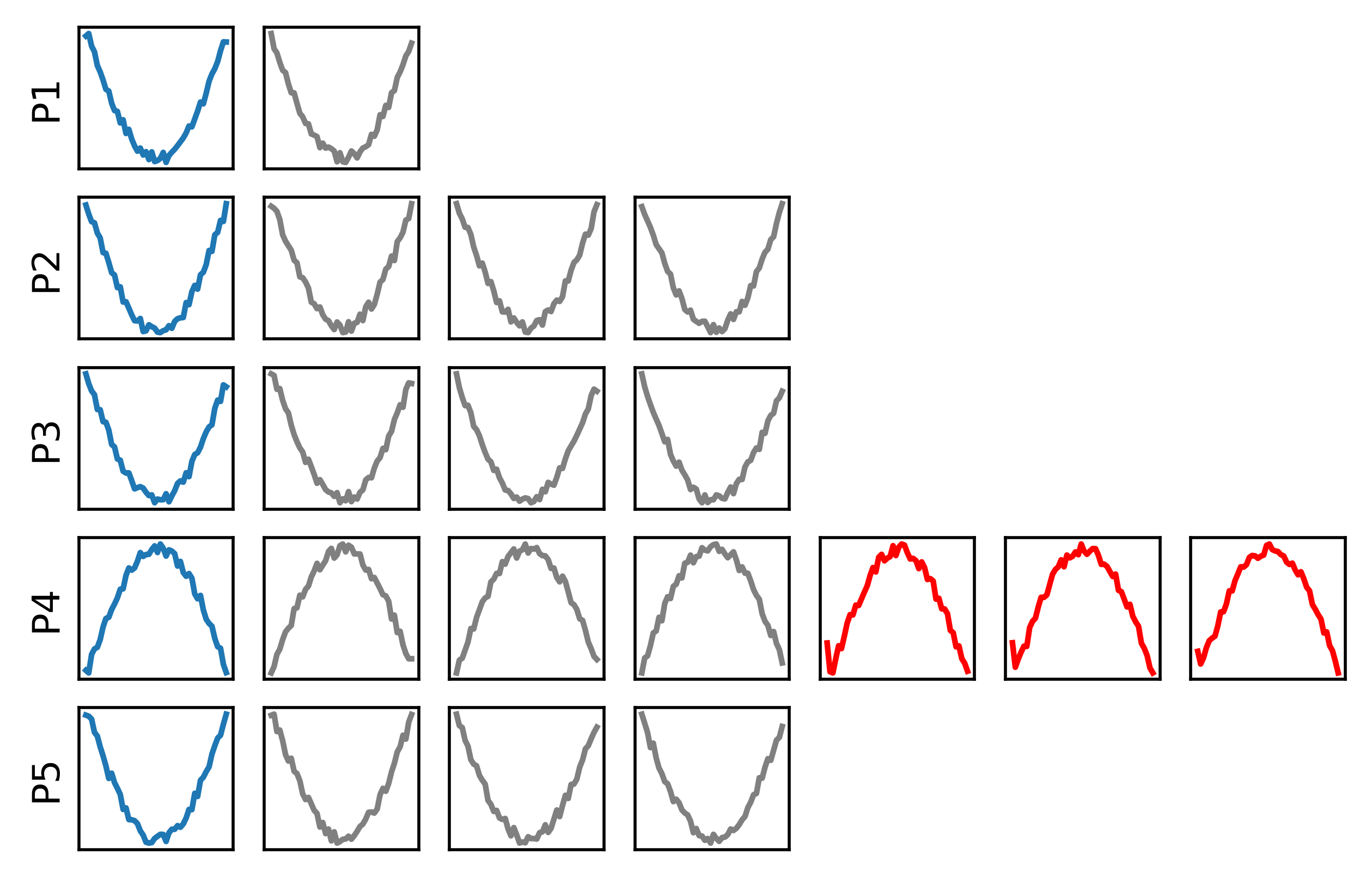}
         \caption{Synthetic regular and anomaly}
         \label{fig:synthtic_na}
     \end{subfigure}
        \caption{Prototype visualization (blue) with assigned regular (grey) and anomaly sequences (red)}
        \label{fig:prototypes_visual}
\end{figure}

\subsubsection{Prototype-based explanation}
Finally, we map the prototypes learned in the latent space back to the human-interpretable input space. Similar to  \cite{chen2018looks,ming2019interpretable}, we map the prototypes back to the input space using the nearest training data embedding in the latent space to prevent unrealistic produced by the decoder. Moreover, each neighbor can only be used once, so every prototype is unique. We visualize the prototypes learned in the one-dimensional datasets Taxi and Synthetic in \autoref{fig:prototypes_visual} with five prototypes ($P1$ to $P5$) for each dataset.
\par
In \autoref{fig:taxi_n} and \autoref{fig:taxi_a}, four similar prototypes ($P1$, $P2$, $P4$, $P5$) show a increasing taxi usage pattern in the morning and turning down at night. $P3$ can be seen as a delayed version of the other four, which is a weekend pattern. The light lines in the background are the regular (grey) and anomaly (red) sequences with the smallest distance to the corresponding prototypes in the latent space. Most of the regular patterns fit the assigned prototypes. A considerable number of both regular and anomaly sequences have the smallest latent space distance to $P3$. Some of them visually fit better with other prototypes, while the distance comparison and prototype assignment do not directly take place in the input space but in latent space. However, this is effective for long and high-dimensional sequences. \autoref{fig:taxi_a} depicts the explanation of anomaly patterns, namely how different are the anomaly sequences to their nearest prototype. 
\autoref{fig:synthtic_na} shows the three assigned regular and anomaly sequences (if available) to each prototype. Since the point anomalies are always generated at the beginning of the crest half, all anomalies are assigned to crest prototype $P4$. \par  
For the high dimensional datasets, we stay observing the prototypes in the latent space and leave the searching for informative sub-input space prototypes as future work. Similarly, we also plan to investigate the reduction of redundant prototypes (e.g., $P1$, $P2$, $P3$, $P5$ in \autoref{fig:synthtic_na}).

\label{sec: latent}

\subsubsection{Efficiency comparison}

\begin{figure}[h]
  \includegraphics[scale=0.4]{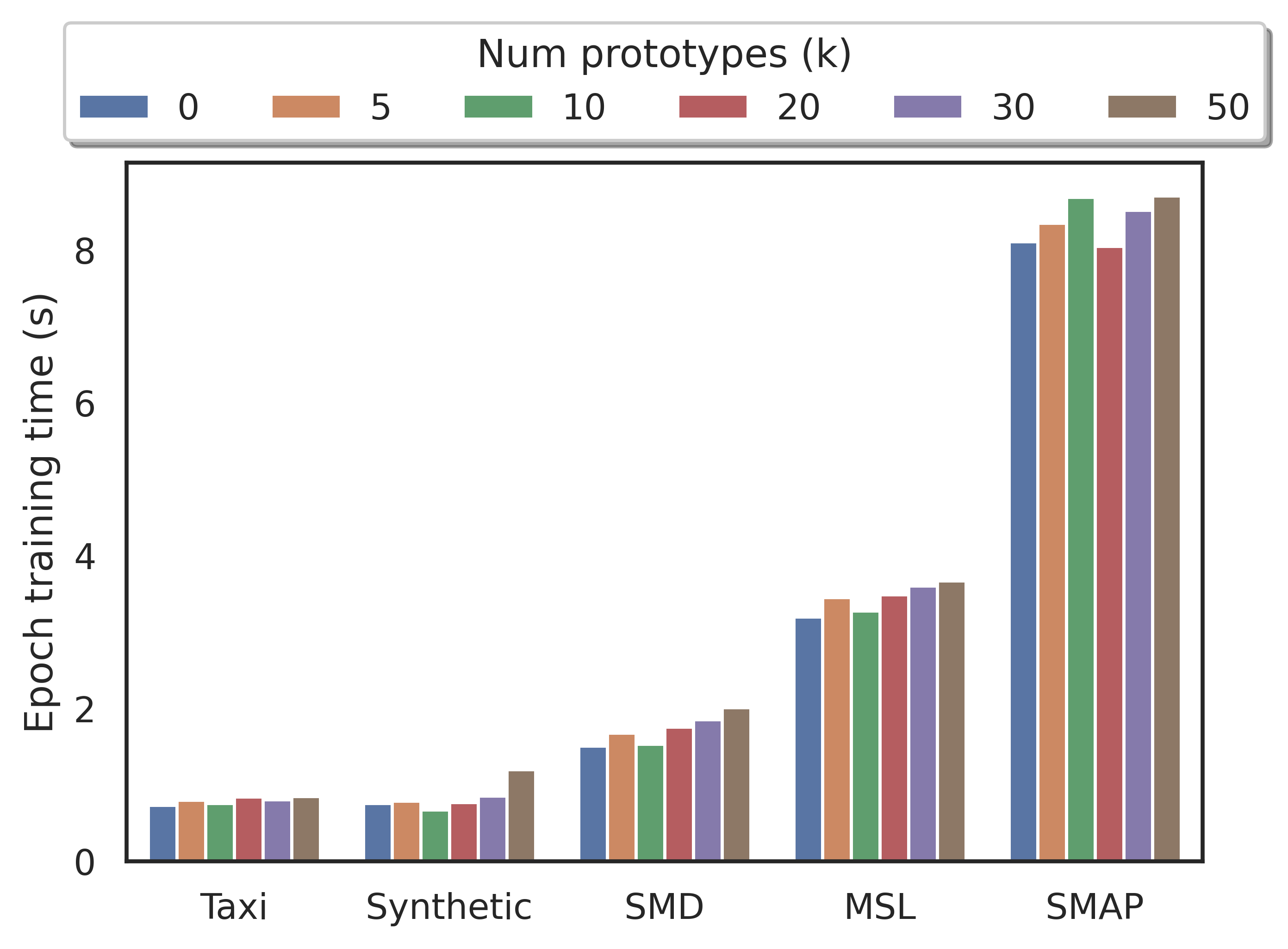}
 \caption{Efficiency analysis}
 \label{fig:time}
\end{figure}

Training the autoencoder with an extra prototype layer does not bring much training expense. We compare the epoch training time between EncDecAD ($k=0$) and ProtoAD ($k\in[5, 10, 20, 30, 50]$) in \autoref{fig:time}. As shown in the figure, there is no significant increase in training time for ProtoAD. 
In the contrary, due to the complex model structure, the epoch training time for OmniAnomaly is: Synthetic $32s$, Taxi $39s$, SMD $225s$, MSL $888s$ and SMAP $3627s$. 

\section{Conclusion and discussion}
In this paper, we explored using prototypes to explain the reconstruction-based anomaly detection process. Specifically, we integrate the recent end-to-end prototype learning into the LSTM-Autoencoder. We use the latent space representations in the autoencoder, which are not directly used in the conventional reconstruction-based anomaly detection models. In our empirical evaluation, we figured out that adding a prototype learning step during the training of the autoencoders does not damage the performance of the autoencoder. The prototypes contribute to an intuitive understanding of the regular pattern state. 
\par
Although the prototypes learned in the two one-dimensional datasets are realistic and interpretable for humans, there are still two major problems to be solved. Firstly, the selection of parameter $k$ is tricky. Pruning techniques can be applied to reduce the redundancy in the prototypes. Moreover, the prototypes of high-dimensional data can only be shown as regular state patterns. However, it is not intuitive enough for humans to directly figure out a small subset of dimensions of interest. In future work, we plan to investigate learning prototypes in subspaces. 

% \newpage
\newpage

\bibliographystyle{ACM-Reference-Format}
\bibliography{main}

\end{document}

%% file: flowchart_new.tex
\begin{figure}[h]
\centering

\tikzset{every picture/.style={line width=0.75pt}} %set default line width to 0.75pt        

\begin{tikzpicture}[x=0.75pt,y=0.75pt,yscale=-0.8,xscale=0.8]
%uncomment if require: \path (0,428); %set diagram left start at 0, and has height of 428

%Shape: Ellipse [id:dp24097469192527143] 
\draw   (270,209.75) .. controls (270,186.69) and (302.94,168) .. (343.58,168) .. controls (384.22,168) and (417.17,186.69) .. (417.17,209.75) .. controls (417.17,232.81) and (384.22,251.5) .. (343.58,251.5) .. controls (302.94,251.5) and (270,232.81) .. (270,209.75) -- cycle ;
%Shape: Star [id:dp7834431038358587] 
\draw  [color={rgb, 255:red, 255; green, 255; blue, 255 }  ,draw opacity=1 ][fill={rgb, 255:red, 245; green, 166; blue, 35 }  ,fill opacity=1 ] (303.99,184.47) -- (306.12,189.29) -- (310.87,190.06) -- (307.43,193.81) -- (308.24,199.11) -- (303.99,196.61) -- (299.73,199.11) -- (300.55,193.81) -- (297.11,190.06) -- (301.86,189.29) -- cycle ;
%Shape: Star [id:dp5183345572239808] 
\draw  [color={rgb, 255:red, 255; green, 255; blue, 255 }  ,draw opacity=1 ][fill={rgb, 255:red, 184; green, 233; blue, 134 }  ,fill opacity=1 ] (397.67,202.23) -- (399.8,207.04) -- (404.56,207.82) -- (401.11,211.57) -- (401.93,216.86) -- (397.67,214.36) -- (393.42,216.86) -- (394.23,211.57) -- (390.79,207.82) -- (395.54,207.04) -- cycle ;
%Shape: Star [id:dp23288659016063507] 
\draw  [color={rgb, 255:red, 255; green, 255; blue, 255 }  ,draw opacity=1 ][fill={rgb, 255:red, 74; green, 144; blue, 226 }  ,fill opacity=1 ] (363.03,226.42) -- (365.16,231.24) -- (369.92,232.01) -- (366.48,235.76) -- (367.29,241.06) -- (363.03,238.56) -- (358.78,241.06) -- (359.59,235.76) -- (356.15,232.01) -- (360.91,231.24) -- cycle ;
%Shape: Ellipse [id:dp6446306200738382] 
\draw   (295.85,182.9) .. controls (295.85,181.19) and (297.1,179.8) .. (298.64,179.8) .. controls (300.19,179.8) and (301.44,181.19) .. (301.44,182.9) .. controls (301.44,184.61) and (300.19,186) .. (298.64,186) .. controls (297.1,186) and (295.85,184.61) .. (295.85,182.9) -- cycle ;
%Shape: Ellipse [id:dp7724490955503223] 
\draw   (307.57,202.37) .. controls (307.57,200.66) and (308.82,199.28) .. (310.37,199.28) .. controls (311.91,199.28) and (313.16,200.66) .. (313.16,202.37) .. controls (313.16,204.08) and (311.91,205.47) .. (310.37,205.47) .. controls (308.82,205.47) and (307.57,204.08) .. (307.57,202.37) -- cycle ;
%Shape: Ellipse [id:dp5499872415568641] 
\draw   (387.5,198.24) .. controls (387.5,196.53) and (388.75,195.14) .. (390.3,195.14) .. controls (391.85,195.14) and (393.1,196.53) .. (393.1,198.24) .. controls (393.1,199.95) and (391.85,201.34) .. (390.3,201.34) .. controls (388.75,201.34) and (387.5,199.95) .. (387.5,198.24) -- cycle ;
%Shape: Ellipse [id:dp7714793412859405] 
\draw   (308.37,183.49) .. controls (308.37,181.78) and (309.62,180.39) .. (311.17,180.39) .. controls (312.71,180.39) and (313.96,181.78) .. (313.96,183.49) .. controls (313.96,185.2) and (312.71,186.59) .. (311.17,186.59) .. controls (309.62,186.59) and (308.37,185.2) .. (308.37,183.49) -- cycle ;
%Shape: Ellipse [id:dp14609699588096448] 
\draw   (296.38,200.6) .. controls (296.38,198.89) and (297.63,197.51) .. (299.18,197.51) .. controls (300.72,197.51) and (301.97,198.89) .. (301.97,200.6) .. controls (301.97,202.31) and (300.72,203.7) .. (299.18,203.7) .. controls (297.63,203.7) and (296.38,202.31) .. (296.38,200.6) -- cycle ;
%Shape: Ellipse [id:dp9707224321885921] 
\draw   (287.32,194.11) .. controls (287.32,192.4) and (288.57,191.01) .. (290.12,191.01) .. controls (291.66,191.01) and (292.91,192.4) .. (292.91,194.11) .. controls (292.91,195.82) and (291.66,197.21) .. (290.12,197.21) .. controls (288.57,197.21) and (287.32,195.82) .. (287.32,194.11) -- cycle ;
%Shape: Ellipse [id:dp29069442737495754] 
\draw   (364.32,220.63) .. controls (365.69,219.85) and (367.38,220.45) .. (368.08,221.98) .. controls (368.79,223.5) and (368.24,225.37) .. (366.87,226.15) .. controls (365.49,226.93) and (363.81,226.32) .. (363.1,224.8) .. controls (362.4,223.28) and (362.94,221.41) .. (364.32,220.63) -- cycle ;
%Shape: Ellipse [id:dp5575023212479205] 
\draw   (362.35,242.42) .. controls (363.73,241.64) and (365.42,242.24) .. (366.12,243.76) .. controls (366.82,245.28) and (366.28,247.15) .. (364.9,247.93) .. controls (363.53,248.71) and (361.84,248.11) .. (361.14,246.59) .. controls (360.44,245.06) and (360.98,243.2) .. (362.35,242.42) -- cycle ;
%Shape: Ellipse [id:dp6737054419858026] 
\draw   (374.18,221.42) .. controls (375.56,220.64) and (377.24,221.24) .. (377.95,222.77) .. controls (378.65,224.29) and (378.11,226.15) .. (376.73,226.93) .. controls (375.36,227.71) and (373.67,227.11) .. (372.97,225.59) .. controls (372.26,224.07) and (372.81,222.2) .. (374.18,221.42) -- cycle ;
%Shape: Ellipse [id:dp666421671495712] 
\draw   (373.13,236.52) .. controls (374.51,235.74) and (376.19,236.34) .. (376.9,237.86) .. controls (377.6,239.39) and (377.06,241.25) .. (375.68,242.03) .. controls (374.31,242.81) and (372.62,242.21) .. (371.92,240.69) .. controls (371.21,239.16) and (371.76,237.3) .. (373.13,236.52) -- cycle ;
%Shape: Ellipse [id:dp4788408587439792] 
\draw   (356.01,224.68) .. controls (357.39,223.9) and (359.07,224.5) .. (359.78,226.02) .. controls (360.48,227.55) and (359.94,229.41) .. (358.56,230.19) .. controls (357.19,230.97) and (355.5,230.37) .. (354.8,228.85) .. controls (354.09,227.33) and (354.64,225.46) .. (356.01,224.68) -- cycle ;
%Shape: Ellipse [id:dp9718145217682346] 
\draw   (351.38,240.56) .. controls (352.76,239.78) and (354.44,240.39) .. (355.15,241.91) .. controls (355.85,243.43) and (355.31,245.3) .. (353.93,246.08) .. controls (352.56,246.86) and (350.87,246.26) .. (350.17,244.73) .. controls (349.46,243.21) and (350.01,241.34) .. (351.38,240.56) -- cycle ;
%Shape: Ellipse [id:dp6950909845792412] 
\draw  [fill={rgb, 255:red, 208; green, 2; blue, 27 }  ,fill opacity=1 ] (349.67,184.67) .. controls (349.67,182.96) and (350.92,181.57) .. (352.46,181.57) .. controls (354.01,181.57) and (355.26,182.96) .. (355.26,184.67) .. controls (355.26,186.38) and (354.01,187.77) .. (352.46,187.77) .. controls (350.92,187.77) and (349.67,186.38) .. (349.67,184.67) -- cycle ;
%Shape: Ellipse [id:dp015367559417519105] 
\draw   (398.69,199.42) .. controls (398.69,197.71) and (399.95,196.33) .. (401.49,196.33) .. controls (403.04,196.33) and (404.29,197.71) .. (404.29,199.42) .. controls (404.29,201.13) and (403.04,202.52) .. (401.49,202.52) .. controls (399.95,202.52) and (398.69,201.13) .. (398.69,199.42) -- cycle ;
%Shape: Ellipse [id:dp9879457235076151] 
\draw   (404.96,213.69) .. controls (404.96,212.08) and (406.14,210.77) .. (407.6,210.77) .. controls (409.06,210.77) and (410.24,212.08) .. (410.24,213.69) .. controls (410.24,215.31) and (409.06,216.62) .. (407.6,216.62) .. controls (406.14,216.62) and (404.96,215.31) .. (404.96,213.69) -- cycle ;
%Shape: Ellipse [id:dp16331820989165047] 
\draw  [fill={rgb, 255:red, 208; green, 2; blue, 27 }  ,fill opacity=1 ] (360.32,196.47) .. controls (360.32,194.76) and (361.58,193.37) .. (363.12,193.37) .. controls (364.67,193.37) and (365.92,194.76) .. (365.92,196.47) .. controls (365.92,198.18) and (364.67,199.57) .. (363.12,199.57) .. controls (361.58,199.57) and (360.32,198.18) .. (360.32,196.47) -- cycle ;
%Shape: Rectangle [id:dp6719526108557901] 
\draw   (309,91) -- (379,91) -- (379,131) -- (309,131) -- cycle ;
%Shape: Rectangle [id:dp5786244613135139] 
\draw   (309,288) -- (379,288) -- (379,328) -- (309,328) -- cycle ;
%Straight Lines [id:da006901731550070789] 
\draw    (343.58,131.5) -- (343.58,166) ;
\draw [shift={(343.58,168)}, rotate = 270] [color={rgb, 255:red, 0; green, 0; blue, 0 }  ][line width=0.75]    (10.93,-3.29) .. controls (6.95,-1.4) and (3.31,-0.3) .. (0,0) .. controls (3.31,0.3) and (6.95,1.4) .. (10.93,3.29)   ;
%Straight Lines [id:da007763982733332897] 
\draw    (343.58,251.5) -- (343.58,286) ;
\draw [shift={(343.58,288)}, rotate = 270] [color={rgb, 255:red, 0; green, 0; blue, 0 }  ][line width=0.75]    (10.93,-3.29) .. controls (6.95,-1.4) and (3.31,-0.3) .. (0,0) .. controls (3.31,0.3) and (6.95,1.4) .. (10.93,3.29)   ;
%Straight Lines [id:da03466147950733889] 
\draw    (344,68.5) -- (344,89.5) ;
\draw [shift={(344,91.5)}, rotate = 270] [color={rgb, 255:red, 0; green, 0; blue, 0 }  ][line width=0.75]    (10.93,-3.29) .. controls (6.95,-1.4) and (3.31,-0.3) .. (0,0) .. controls (3.31,0.3) and (6.95,1.4) .. (10.93,3.29)   ;
%Straight Lines [id:da832998684629586] 
\draw    (343.58,327.25) -- (343.58,348.25) ;
\draw [shift={(343.58,350.25)}, rotate = 270] [color={rgb, 255:red, 0; green, 0; blue, 0 }  ][line width=0.75]    (10.93,-3.29) .. controls (6.95,-1.4) and (3.31,-0.3) .. (0,0) .. controls (3.31,0.3) and (6.95,1.4) .. (10.93,3.29)   ;
%Straight Lines [id:da2507519788390221] 
\draw    (474.17,62.5) -- (364.17,62.5) ;
%Straight Lines [id:da8542760292575905] 
\draw    (474.17,363.5) -- (364.17,363.5) ;
%Straight Lines [id:da42237618823920353] 
\draw    (474.17,62.5) -- (474.17,191.5) ;
\draw [shift={(474.17,193.5)}, rotate = 270] [color={rgb, 255:red, 0; green, 0; blue, 0 }  ][line width=0.75]    (10.93,-3.29) .. controls (6.95,-1.4) and (3.31,-0.3) .. (0,0) .. controls (3.31,0.3) and (6.95,1.4) .. (10.93,3.29)   ;
%Straight Lines [id:da15285984927282636] 
\draw    (474.17,363.5) -- (474.17,232.5) ;
\draw [shift={(474.17,230.5)}, rotate = 90] [color={rgb, 255:red, 0; green, 0; blue, 0 }  ][line width=0.75]    (10.93,-3.29) .. controls (6.95,-1.4) and (3.31,-0.3) .. (0,0) .. controls (3.31,0.3) and (6.95,1.4) .. (10.93,3.29)   ;
%Straight Lines [id:da06173848098396484] 
\draw    (528.17,216) -- (555,216) ;
\draw [shift={(557,216)}, rotate = 180] [color={rgb, 255:red, 0; green, 0; blue, 0 }  ][line width=0.75]    (10.93,-3.29) .. controls (6.95,-1.4) and (3.31,-0.3) .. (0,0) .. controls (3.31,0.3) and (6.95,1.4) .. (10.93,3.29)   ;

% Text Node
\draw (314,101) node [anchor=north west][inner sep=0.75pt]   [align=left] {Encoder};
% Text Node
\draw (313,296) node [anchor=north west][inner sep=0.75pt]   [align=left] {Decoder};
% Text Node
\draw (328,46) node [anchor=north west][inner sep=0.75pt]    {$W_{t}$};
% Text Node
\draw (330,350) node [anchor=north west][inner sep=0.75pt]    {$W_{t}^{'}$};
% Text Node
\draw (559,191) node [anchor=north west][inner sep=0.75pt]    {$ \begin{array}{l}
Anomaly\\
\ \ \ score
\end{array}$};
% Text Node
\draw (422,191) node [anchor=north west][inner sep=0.75pt]    {$ \begin{array}{l}
Reconstruction\\
\ \ \ \ \ \ \ \ error
\end{array}$};

\end{tikzpicture}
\caption{ProtoAD overview: ProtoAD consists of a encoder and a decoder, which reconstruct the input data window $W_t$ to $W^{'}_t$. The anomaly score is calculated based on the reconstruction error. In the latent space, prototypes (stars) are learned as representative points of the encoded normal windows (white dots). In the test phase, latent representations of the abnormal data windows (red dots) can also be explained by comparing with the learned prototypes.}
\label{fig:overview}
\end{figure}